\pdfoutput=1

\documentclass[11pt]{article}

\usepackage[]{emnlp2021}

\usepackage{times}
\usepackage{latexsym}
\usepackage{pgf-pie}

\usepackage[T1]{fontenc}

\usepackage[utf8]{inputenc}

\usepackage{microtype}

\usepackage{booktabs}
\usepackage{multirow}
\usepackage{graphicx}
\usepackage{soul}
\usepackage{mwe}
\usepackage{subfig}

\newcommand{\repolink}[0]{www.github.com/SLAB-NLP/BUG}
\newcommand{\spikeurl}[0]{spike.apps.allenai.org}
\newcommand{\spiketeam}[0]{Micah Shlain, Hillel Taub-Tabib, Shoval Sadde, and Yoav Goldberg}

\title{Collecting a Large-Scale Gender
Bias Dataset for Coreference Resolution and Machine Translation}

\author{
 Shahar Levy \qquad
 Koren Lazar \qquad
 Gabriel Stanovsky\\
 School of Computer Science and Engineering\\
 The Hebrew University of Jerusalem,  Jerusalem, Israel \\
 shaharl6000@gmail.com \quad 
 \{koren.lazar, gabriel.stanovsky\}@mail.huji.ac.il
}

\begin{document}
\maketitle
\begin{abstract}
Recent works have found evidence of gender bias in models of machine translation and coreference resolution using mostly synthetic diagnostic datasets. While these quantify bias in a controlled experiment, they often do so on a small scale and consist mostly of artificial, out-of-distribution sentences.
In this work, we find grammatical patterns indicating stereotypical and non-stereotypical gender-role assignments (e.g., female nurses versus male dancers) in corpora from three domains, resulting in a first large-scale gender bias dataset of 108K diverse real-world English sentences.
We manually verify the quality of our corpus and use it to evaluate gender bias in various coreference resolution and machine translation models. We find that all tested models tend to over-rely on gender stereotypes when presented with natural inputs, which may be especially harmful when deployed in commercial systems.
Finally, we show that our dataset lends itself to finetuning a coreference resolution model, finding it mitigates bias on a held out set.
Our dataset and models are publicly available at \url{\repolink}. We hope they will spur future research into gender bias evaluation mitigation techniques in realistic settings.

\end{abstract}


\section{Introduction}
Gender bias in machine learning occurs when supervised models predict based on spurious societal correlations in their training data. This may result in harmful behaviour when it occurs in models deployed in real-world applications~\citep{Caliskan_2017,pmlr-v81-buolamwini18a, 10.1145/3442188.3445922}.\footnote{We acknowledge that gender identity is non-binary. Throughout this work we refer to \emph{grammatical} gender, which has categorical inflections in the discussed languages (e.g., masculine and feminine pronouns in English).}


\begin{figure}[tb!]
    \centering
    \includegraphics[width=\linewidth]{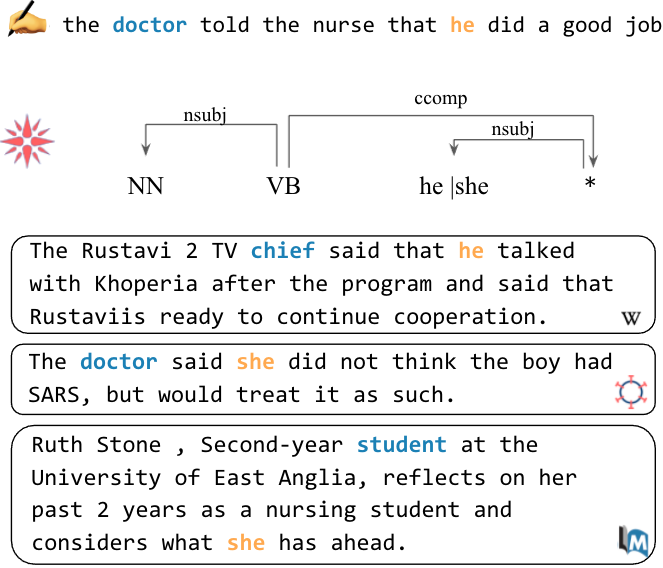}
    
    \caption{ \label{fig:flow}
    We propose a semi-automatic method to vastly extend synthetic, small diagnostic datasets. We start with the texts of Winogender~\citep{rudinger-etal-2018-gender} and WinoBias~\citep{zhao-etal-2018-gender}, specifically designed to to be challenging for coreference and machine translation (top),  extract syntactic patterns focusing on the salient entities in the artificial sentences (middle), and query real-world datasets for matching texts, using SPIKE~\citep{shlain-etal-2020-syntactic}. The result is a large collection of diverse real-world texts exhibiting similar challenging properties which lends itself to both finetuning and testing (bottom).
    }
    
\end{figure}

Recent work has quantified bias mostly using carefully designed templates, following the Winograd schema~\citep{levesque2012winograd}.  \citet{zhao-etal-2018-gender} and \citet{rudinger-etal-2018-gender} probed for gender bias in coreference resolution with templates portraying two human entities and a single pronoun. For example,  given the sentence \emph{``the doctor asked the nurse to help her because she was busy''}, models often erroneously cluster ``her'' with ``nurse'', rather than with ``doctor''. \citet{stanovsky-etal-2019-evaluating} used the same data to evaluate gender bias in machine translation. When translating this sentence to a language with grammatical gender, models tend to inflect nouns based on stereotypes, e.g., in Spanish, preferring the masculine inflection over the correct feminine inflection (\emph{``doctor-a''}).

While these experiments are useful for quantifying gender bias in a controlled environment, we identify two shortcomings with this approach. 
First, the artificially-constructed texts diverge from natural language training distribution, which may inadvertently cause models to use prior distributions on such unseen constructions.
Second, the small-scale templated data does not lend itself to training or finetuning to mitigate gender bias, limiting these datasets to diagnostic purposes.

\begin{figure*}[tb!]
    \centering
    \includegraphics[width=\linewidth]{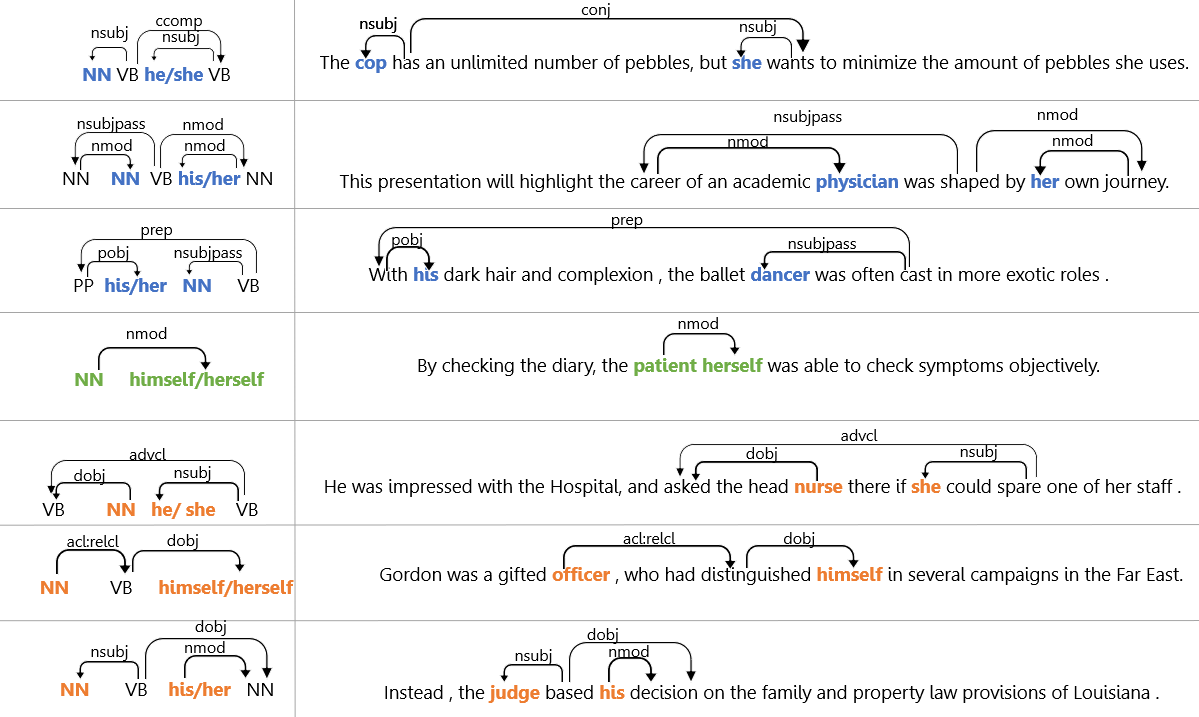}
    \caption{Grammatical patterns (left) and corresponding examples sentences from BUG (right).
    Each instance in our dataset ensures at least a single human entity (marked by their profession) and a gendered pronoun, marked in bold. The sentences marked in blue are classified as anti-stereotypical while the sentences marked in orange are classified as stereotypical, and the sentence marked in green classified is neutral. The figure depicts 7 templates out of the 14 we designed. See the Appendix for a complete list.
    } 
    \label{fig:examples}
\end{figure*}

In this work, outlined in Figure~\ref{fig:flow}, we address both of these limitations by creating BUG, a large-scale dataset of 108K sentences, sampled semi-automatically from large corpora using lexical-syntactic pattern matching~(see Figure~\ref{fig:examples} for examples). To construct BUG, we devise 14 diverse syntactic patterns, matching a wide range of sentences, ensuring that each mentions a human entity and a pronoun referring to it. Following, we use the SPIKE engine~\cite{shlain-etal-2020-syntactic}\footnote{\url{\spikeurl}} to retrieve matching sentences over three diverse domains, including Wikipedia, Covid19 research, and PubMed abstracts.  Finally, we filter the resulting sentences and mark each as either stereotypical or anti-stereotypical with respect to gender role assignments. The result is large corpus which is diverse, challenging, and accurate.

We use BUG to conduct a first large-scale evaluation of gender bias on real-world texts.
We find that popular machine translation and coreference models struggle with feminine entities and anti-stereotypical assignments. Furthermore, BUG enables us to identify novel insights. For example, that machine translation models tend to be more biased when there are many pronouns in the input sentence.

Finally, we show that BUG can also help in mitigating gender bias. We finetune a state-of-the-art coreference resolution model on the anti-stereotypical portion of BUG and achieve a 50\% error reduction on a held out test set, at the cost of only a modest drop in overall accuracy. 

To conclude, our main contributions are:

\begin{itemize}
    \item We present BUG, a first publicly available large-scale corpus for gender bias evaluation which consists of diverse, real-world sentences.
    
    \item We evaluate gender bias at large scale on natural sentences, leading to novel insights in machine translation and coreference resolution.
    
    \item We use BUG to finetune a coreference resolution model, showing that the resulting model is less prone to make gender biased predictions.
\end{itemize}

\section{Data Collection}
In this section, present BUG, a semi-automatic collection of natural, ``in the wild'' English sentences which are challenging with respect to societal gender-role assignments.
Similarly to some of the synthetic gender bias datasets~\citep{zhao-etal-2018-gender,rudinger-etal-2018-gender}, we are looking for sentences with a human entity, identified by their profession (e.g., ``cop'', ``dancer'') and a gendered pronoun (e.g., ``he'', ``she''). For example, see the first sentence in Figure~\ref{fig:examples}, where the cop co-refers with a feminine pronoun (``she''), while the judge in the last sentence in Figure~\ref{fig:examples} co-refers with a masculine pronoun (``his''). 

As opposed to previous work, we are interested in naturally occurring sentences, rather than generating artificial sentences from fixed lexical templates.
The process for achieving this is outlined in Figure~\ref{fig:flow} and elaborated below.
First, we perform syntactic search for sentences with challenging syntactic properties over corpora from three domains~(Section~\ref{sec:collecting}). We then filter the sentences to verify they contain at least one entity, and a corresponding pronoun~(Section~\ref{sec:filter-profession}). Finally, we manually assess BUG, finding it to be 85\% accurate~(Section~\ref{sec:human}).

\subsection{Syntactic Querying with SPIKE}
\label{sec:collecting}
We devised 14 lexical-syntactic patterns, exemplified in Figure~\ref{fig:examples} to construct BUG. 
All our patterns have two anchors --- a pronoun and a profession --- which the pattern indicates are coreferring.\footnote{See Appendix for a full list of patterns: \url{github.com/SLAB-NLP/BUG/blob/main/docs/appendix.pdf}}

For example, the last pattern in the figure links a noun (e.g., ``officer'') with a relative clause relation (``acl:relcl'') to a verb (e.g., ``distinguished'') modified by a direct object (``dobj'') gendered reflexive pronoun (``himself'' or ``herself'').
These patterns were constructed by examining and expanding the sentences in the synthetic coreference corpora~\cite{rudinger-etal-2018-gender,zhao-etal-2018-gender}. 

To match these 14 patterns against real-world texts, we used SPIKE \cite{shlain-etal-2020-syntactic}, which indexes large-scale corpora and retrieves matching instances given a lexical-syntactic pattern.
We queried corpora from three domains: Wikipedia, PubMed abstracts, and Covid19 research papers~\citep{wang-etal-2020-cord}.
The examples in Figure~\ref{fig:examples} highlight the diversity of the approach, while they all adhere to one of the predefined patterns, they  vary widely in vocabulary and in syntactic construction, often introducing complex phenomena, such as coordination or adverbial phrases.



\subsection{Marking Entities and Gender Roles}
\label{sec:filter-profession}
Following the lexical-syntactic querying, we filter BUG to make sure it contains human entities, and mark each instance as either stereotypical (bottom three examples in Figure~\ref{fig:examples}), neutral (middle example) or anti-stereotypical (top three examples). This enables us to use BUG to measure gender bias in machine translation and coreference resolution models~(Section~\ref{sec:wild}).

We filter out two types of nouns:
(1) nouns which do not refer to a person (e.g., ``COVID-19''); (2) gendered English nouns (e.g., ``princess'', ``father'', or ``sister''). To address both of these issues, we filtered the results with a predefined list of 183 professions, taken from the U.S. census.

\begin{table*}[!t]
\small
\centering

\begin{tabular}{  p{\dimexpr 0.17\linewidth-2\tabcolsep} 
                   p{\dimexpr 0.41\linewidth-2\tabcolsep} 
                   p{\dimexpr 0.41\linewidth-2\tabcolsep} } 
                   \toprule
\textbf{Category} & \textbf{Example} & \textbf{Comments} \\\midrule
Disambiguated by noun~(67\%) & A \textbf{{\color[HTML]{FE0000}physician}} who respects \textbf{{\color[HTML]{FE0000}her}} autonomy should respect Ann’s right to make this decision. & Noun selection affects coreference decision. E.g., replacing ``autonomy'' with ``job'' would lead to a correct annotation. \\\hline
Ambiguous (23\%) & Hiei's \textbf{{\color[HTML]{6434FC}captain}} ordered \textbf{{\color[HTML]{6434FC}her}} crew to abandon ship after further damage. & The antecedent is ambigous (either captain or Hiei). \\\hline
Non-gendered pronoun (7\%) &  The IPP is a portfolio in which the \textbf{{\color[HTML]{329A9D}student}} reflects on \textbf{{\color[HTML]{329A9D}his/her}} learning and development during the production. & Reference to masculine and feminine pronouns. \\\hline
Reported speech (3\%) & We remove the comments , but this person keeps putting them back up - things like “\textbf{{\color[HTML]{CD9934}he}} says he never met that woman”. & Quoted pronoun which does not refer to an entity in the sentence.\\
\bottomrule
\end{tabular}             
\caption{Error analysis of 30 errors found in a sample of 200 randomly sampled sentences from BUG.}
\label{tab:ex-errors}
\end{table*}

Following, to mark each instance as either stereotypical or anti-stereotypical, we we follow \citet{zhao-etal-2018-gender} and \citet{rudinger-etal-2018-gender} and use the United States 2015 census'  gender distribution per occupation.\footnote{https://www.kaggle.com/jonavery/incomes-by-career-and-gender} 
For instance, the first example Figure~\ref{fig:examples} is marked anti-stereotypical since ``cop'' is a predominantly male profession (76\% in the census) and the referring pronoun is feminine. 


\subsection{Human Validation and Gold Standard}
\label{sec:human}
We estimate the accuracy of BUG by randomly sampling 1700 sentences from BUG, sampling uniformly across the data as well as from every pattern and domain.
17 human annotators proficient in English were asked to decide whether the gender BUG assigned to the entity matches their understanding of the sentence. The complete annotation guideline is presented in the Appendix.
Overall we found that 85\% of the instances were marked  correct. 
We publish these annotation as a separate resource of diverse sentences with gold annotations (dubbed \emph{Gold BUG}).



\begin{table*}[]
\centering
\captionsetup{justification=raggedright, singlelinecheck=false}
\resizebox{\textwidth}{!}{%
\begin{tabular}{@{}ccccccc@{}}
\toprule
\textbf{}                                                                                                      \textbf{Corpus}  & \textbf{Stereotypical}                             & \textbf{Anti-stereotypical} & \textbf{Neutral}                                   & \textbf{Male}               & \textbf{Female}             & \textbf{Total}                      \\ \midrule
WinoGender + WinoBias                                                                            & \multicolumn{1}{c}{1,584}                         & \multicolumn{1}{c}{1,584}                                            & \multicolumn{1}{c}{720}                    & \multicolumn{1}{c}{1,826}  & \multicolumn{1}{c}{2,062}  & \multicolumn{1}{c}{3,888} \\ 
GAP$^*$                                                                               & \multicolumn{1}{c}{\textbf{-}}                    & \multicolumn{1}{c}{\textbf{-}}                                       & \multicolumn{1}{c}{\textbf{-}}                    & \multicolumn{1}{c}{2,227}  & \multicolumn{1}{c}{2,227}  & \multicolumn{1}{c}{4,454} \\ \midrule \midrule

 Wikipedia                                                  & \multicolumn{1}{c}{{ 48,909}} & \multicolumn{1}{c}{{ 25,529}}                    & \multicolumn{1}{c}{{ 5,607}}  & \multicolumn{1}{c}{63,677} & \multicolumn{1}{c}{16,368} & \multicolumn{1}{c}{80,045}         \\ 
Pubmed abstracts & \multicolumn{1}{c}{{ 4,099}}  & \multicolumn{1}{c}{{ 3,665}}                     & \multicolumn{1}{c}{{ 16,543}} & \multicolumn{1}{c}{16,021} & \multicolumn{1}{c}{8,286}  & \multicolumn{1}{c}{24,307}         \\ 
Covid19 research                       & \multicolumn{1}{c}{1,001}                         & \multicolumn{1}{c}{683}                                              & \multicolumn{1}{c}{2,383}                         & \multicolumn{1}{c}{2,572}  & \multicolumn{1}{c}{1,495}  & \multicolumn{1}{c}{4,067}          \\ 
Balanced BUG                       & 12,922                         & 12,922                                              & -                         & 12,922  & 12,922  & \multicolumn{1}{c}{25,844}          \\
Gold BUG                       & 865                         & 420                                              & 435                         & 1,337  & 383  & \multicolumn{1}{c}{1,720}          \\\midrule
\textbf{BUG Total}                                                                                               & \textbf{54,009}                                             & \textbf{29,877}                                                                & \textbf{24,533}                                             & \textbf{82,270}                      & \textbf{26,149}                      & \textbf{108,419}                    \\ \bottomrule

\end{tabular}
}
\caption{\label{tab:count_sentences} Statistics for existing gender bias datasets (top) versus different BUG subsets (bottom). Stereotypical, anti-stereotypical and neutral refer to societal gender role assignments. E.g., a sentence with male doctor is stereotypical, while a sentence with a female doctor is anti-stereotypical;   male, female refer to the number of sentences with masculine and feminine pronouns. 
BUG contains sentences from the three corpora listed above it. WinoMT contains sentences from WinoGender and Winobias. $^*$Sentences in GAP do not have stereotypical classification. 
}

\end{table*}


\section{BUG Analysis}
\label{sec:data}
The collection described in the previous section resulted in 108k sentences and 1700 human annotations. 
Following, we analyze key characteristics of BUG, finding it to be lexically diverse, and an order of magnitude larger than previous gender bias corpora.

\subsection{Error Analysis and Inter-Annotator Agreement} 

The error analysis in Table~\ref{tab:ex-errors} reveals that the most common errors are due to  constructions where syntactic patterns are ambiguous with respect to coreference.

For instance, in the first example in Table~\ref{tab:ex-errors}, replacing ``autonomy" with ``job''  changes the antecedent from the physician to the patient. Future work may address this by trying to refine our lexical-syntactic patterns to also include verb selection information. 

Other types of errors were less frequent and included cases where two pronouns were used as a single gender-neutral word (``he/she''), and where the pronoun was part of a named entity or reported speech.


In addition, we test agreement between two annotators on a subset of 200 randomly selected sentences. We found a high level of agreement~(95.5\%; 0.73$\kappa$). Disagreements mostly occur on ambiguous sentences, such as 
``On the night of 17 August , Charlotte reported that the \emph{child} had been taken from \emph{her} tent by a dingo .'', where one annotator read ``her'' as referring to the child, while the other thought that the pronoun refers to Charlotte.

\subsection{Data Characteristics}
\label{sec:characteristics}

\begin{figure}[tb!]
    \centering
    \includegraphics[width=\linewidth]{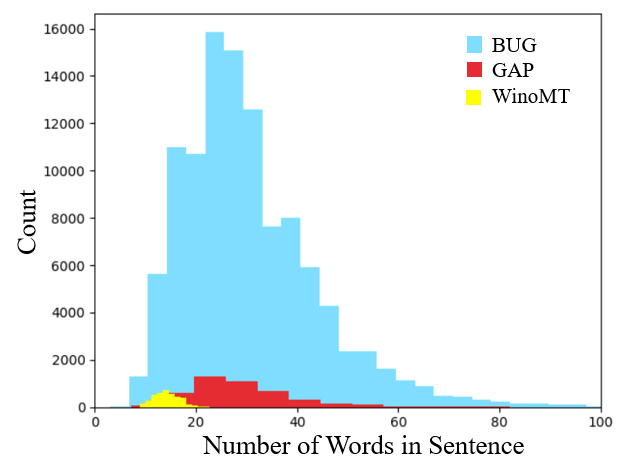}
    
    \caption{The distribution of the number of words in a sentence in BUG (in blue, average - 30.6 words per sentence) versus WinoMT used in \citet{stanovsky-etal-2019-evaluating} (in yellow, average - 14.3 words per sentence) and GAP used in \citet{webster-etal-2018-mind} (in red, average - 29.8 words per sentence). Word splitting was done with spaCy~\citep{spacy}.}
    \label{fig:hist_length}
\end{figure}

\begin{figure}[tb!]
    \centering
    \includegraphics[width=\linewidth]{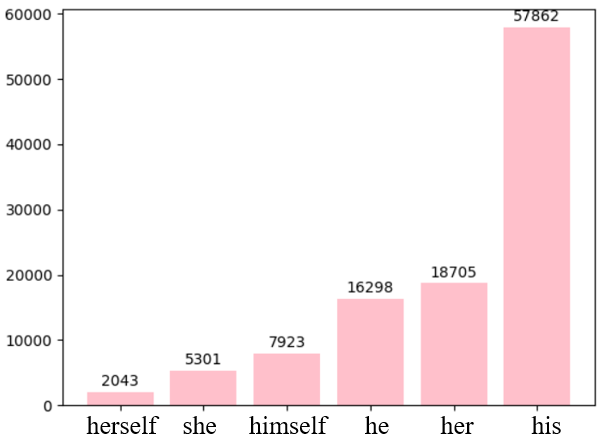}
    
    \caption{\label{fig:pronouns_dist}BUG pronoun histogram. In total, there are 82K  (76\%) masculine pronouns and 26K (24\%) feminine pronouns.
    }
    
\end{figure}

BUG statistics are presented in Table~\ref{tab:count_sentences} in comparison with other datasets for gender bias. 
BUG is more than 24 times larger than the GAP coreference challenge set~\citep{webster-etal-2018-mind} and more than 30 times larger than WinoMT (WinoGender and Winobias combined)~\citep{stanovsky-etal-2019-evaluating}. 
BUG consists of 110,544 unique words, while in the WinoMT corpus the vocabulary size is 1,868 and GAP's vocabulary size is 31,834.
BUG is more diverse and naturally distributed, as can be seen in the histogram of sentence lengths depicted in Figure~\ref{fig:hist_length}. 
Furthermore, the mean distance (in words) between entity and pronoun does not significantly differs between stereotypical ($6.4 [\pm 4.5]$) and anti-stereotypical ($6.3 [\pm 4.6]$) partitions, thus alleviating recent concerns about such artifacts in diagnostic datasets~\citep{Kocijan2021TheGO}.



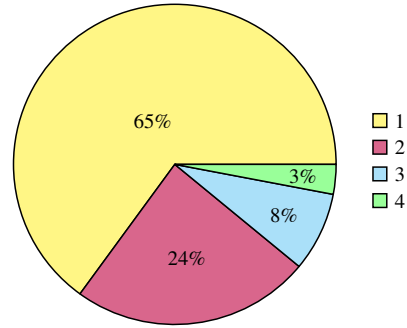
\begin{figure}
\center
 \resizebox{0.7\linewidth}{!}{
\begin{tikzpicture}
 
\pie[text=legend,color={yellow!60, purple!60, cyan!30, green!40}]{
65/1,
24/2,
8/3,
3/4
}

\end{tikzpicture}
}
\caption{\label{fig:pron}The distribution of the number of pronouns in our corpus. 35\% (41K) of the sentences have more than one pronoun, further complicating the coreference resolution task.}
\end{figure}

Our sentences were sampled from three corpora indexed in SPIKE. The majority were drawn from Wikipedia. Relative to the size of the original corpora, the yield from Wikipedia is 6 times more productive than PubMed and 4 times more than the Covid19 research domain. This is possibly since Wikipedia lends itself more to discussion of different entities in different settings.

As expected, since BUG is sampled from real texts, most of the data is stereotypical and most entities are male.
There are three times more sentences with masculine pronouns compared to feminine pronouns, as shown in Figure~\ref{fig:pronouns_dist}; there are twice as many sentences with typically-male professions compared to typically-female professions; and twice as many sentences classified as stereotypical than anti-stereotypical.
The natural texts also present a more challenging coreference setting. As evident in Figure~\ref{fig:pron} by large number of instances (35\% of the corpus) with more than one pronoun.

To allow for more controlled evaluations, we publish two subsets of BUG. \emph{Gold BUG} consists of the gold-quality human-validated samples, while \emph{Balanced BUG} is randomly sampled from BUG to ensure balance between male and female entities and between stereotypical and non-stereotypical gender role assignments. We report statistics for both of these subsets in Table \ref{tab:count_sentences}.

\begin{table*}[tb!]
\centering
\begin{tabular}{@{}l|lll|lll|lll@{}}
\toprule
\multirow{2}{*}{\begin{tabular}[c]{@{}l@{}}Target \\ Language\end{tabular}} & \multicolumn{3}{c|}{Opus-MT}               & \multicolumn{3}{c|}{mBART50\_m2m}               & \multicolumn{3}{c|}{m2m\_100\_418M}       \\ \cmidrule(l){2-10} 
                                                                            & Acc           & $\Delta_G$ & $\Delta_S$ & Acc           & $\Delta_G$ & $\Delta_S$ & Acc  & $\Delta_G$ & $\Delta_S$ \\ \midrule
Arabic                                                                      & 75.4          & 19.1       & 12.4       & \textbf{79.5} & 26.0       & 15.5       & 73.8 & 52.3       & 16.6       \\
Czech                                                                       & 83.2          & 26.3       & 23.6       & \textbf{85.0} & 20.7       & 21.7       & 76.1 & 48.6       & 20.4       \\
German                                                                      & 75.7          & 24.9       & 15.2       & \textbf{77.2} & 25.0       & 17.2       & 70.0 & 44.1       & 16.8       \\
Spanish                                                                     & \textbf{63.4} & 20.5       & 15.5       & 62.8          & 20.1       & 15.8       & 57.1 & 43.9       & 15.5       \\
Hebrew                                                                      & \textbf{75.8} & 28.4       & 24.3       & 57.7          & 14.8       & 21.2       & 73.3 & 45.9       & 20.3       \\
Italian                                                                     & 58.8          & 32.8       & 19.8       & \textbf{61.1} & 27.2       & 20.9       & 55.8 & 48.6       & 20.8       \\
Russian                                                                     & 68.7          & 47.1       & 17.3       & \textbf{73.5} & 33.4       & 12.6       & 68.6 & 55.2       & 13.9       \\
Ukrainian                                                                   & 67.1          & 35.4       & 17.3       & \textbf{71.5} & 26.1       & 16.2       & 67.8 & 48.4       & 15.8       \\ \bottomrule
\end{tabular}
\caption{Results for machine translation gender bias evaluation evaluation across 8 diverse target languages on the BUG dataset.
$Acc$ represents the overall accuracy (F1) of gender translation. $\Delta_G$ is the difference in accuracy between masculine and feminine entities. $\Delta_S$ is the difference in performance between stereotypical and anti-stereotypical gender role assignments.
Positive $\Delta_G$ and $\Delta_S$ values indicate that the translations are gender biased.}
\label{tab:mt-results}
\end{table*}

\section{Evaluating Gender Bias in The Wild}
\label{sec:wild}

We evaluate the performance of machine translation and coreference resolution models on BUG, using the metrics and tools established in  previous work~\citep{rudinger-etal-2018-gender,zhao-etal-2018-gender,stanovsky-etal-2019-evaluating}.
To the best of our knowledge, this is the first quantitative evaluation of gender bias in such systems on a large scale using naturally occurring sentences.
Such inputs better resemble real-world use where biases can affect many users.


\subsection{Experimental Setup}

\paragraph{Machine translation.}
We used EasyNMT\footnote{\url{https://github.com/UKPLab/EasyNMT}} to evaluate three machine translation models:  mBART50\_m2m~\cite{tang2020multilingual,liu-etal-2020-multilingual-denoising}, m2m\_100\_418M~\cite{fan2020englishcentric}, and Opus-MT~\citep{TiedemannThottingal:EAMT2020}, representing the state-of-the-art for publicly available neural machine translations models.
We translated BUG from English to a set of eight diverse target languages with grammatical gender: Arabic, Czech, German, Spanish, Hebrew, Italian, Russian and Ukrainian, using tools developed in previous work to infer the translated gender based on morphological inflections~\citep{stanovsky-etal-2019-evaluating,kocmi-etal-2020-gender}.\footnote{We used the implementation provided by \url{github.com/gabrielStanovsky/mt_gender}}

\paragraph{Coreference resolution.}
We use the AllenNLP~\citep{gardner-etal-2018-allennlp} implementation of SpanBERT~\citep{joshi-etal-2020-spanbert}.
SpanBERT introduces contextual span representation to the the e2e-coreference model~\citep{lee-etal-2018-higher,joshi-etal-2019-bert} to achieve state-of-the-art results on the English portion of the popular CoNLL-2012 shared task coreference benchmark~\citep{pradhan-etal-2012-conll}.

\subsection{Metrics}
For each tested model we compute three metrics, following \citet{zhao-etal-2018-gender} and \citet{stanovsky-etal-2019-evaluating}, while adapting the terminology suggested recently by \citet{mehrabi2021}.

\paragraph{Accuracy:}
Denotes the F1 score of the gender prediction.
For machine translation, this indicates the percentage of instances in which a correct grammatical gender inflection was produced in the target language. For example translating a female doctor as \emph{doctor-a} in Spanish. For coreference resolution accuracy refers to the portion of instances where the entity's antecedent is correctly clustered with its pronoun, e.g., a female doctor clustered with the feminine pronoun ``her''.

\begin{figure*}[tb!]

\begin{minipage}{.33\linewidth}
\centering
\subfloat[Accuracy]{\label{main:a}\includegraphics[width=\textwidth]{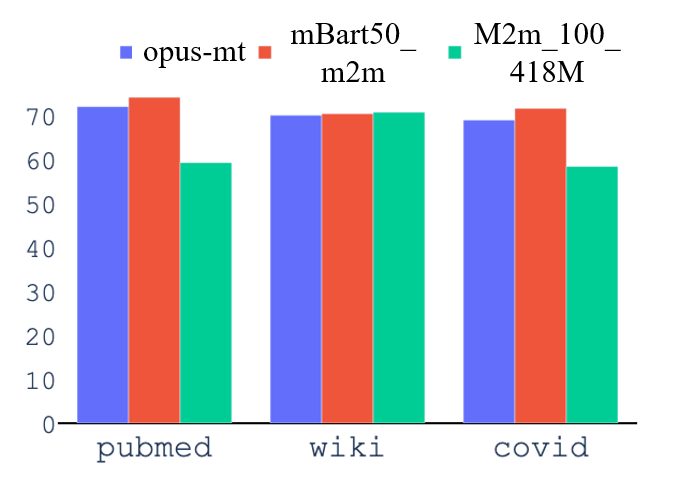}}
\end{minipage}\hfill%
\begin{minipage}{.33\linewidth}
\subfloat[$\Delta_G$]{\label{main:a}\includegraphics[width=\textwidth]{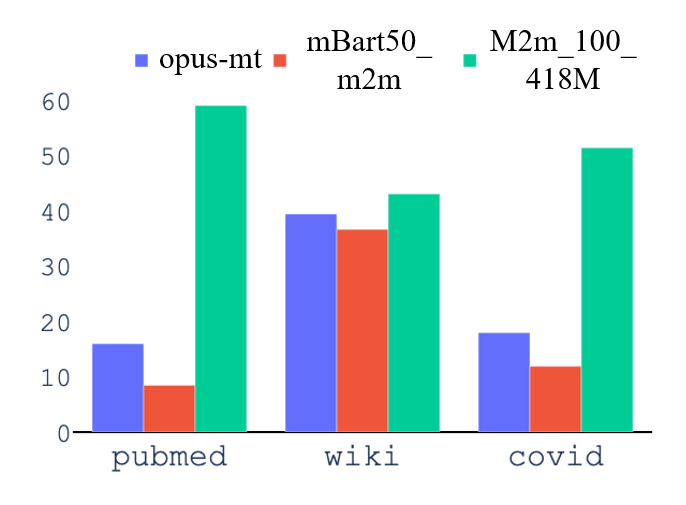}}
\end{minipage}\hfill%
\begin{minipage}{.33\linewidth}
\subfloat[$\Delta_S$]{\label{main:a}\includegraphics[width=\textwidth]{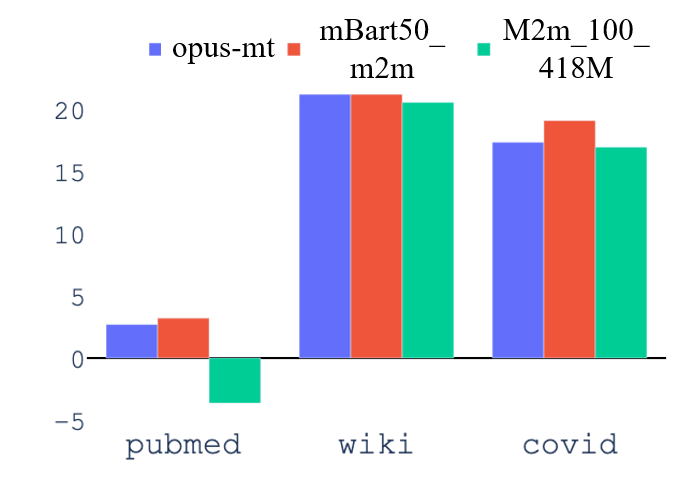}}
\end{minipage}%
\caption{Gender bias in machine translation across the domains in BUG.}
\label{fig:main}
\end{figure*}

\paragraph{Population bias ($\Delta_G$):\footnote{According to \citet{mehrabi2021}, \emph{Population bias}  occurs when ``statistics, demographics, representatives, and
user characteristics are different in the user population represented in the dataset or platform
from the original target population''~\citep{Olteanu2019}.}} denotes the difference in accuracy (F1 score) between sentences with entities which co-refer with a masculine pronoun versus those with entities which co-refer with feminine pronouns.
By definition, $-100 \geq \Delta_G \geq 100$. When $\Delta_G > 0$, the model tends to perform better when the input entities co-refer with masculine pronouns, and conversely when $\Delta_G < 0$ it performs better when they co-refer with feminine ones.

\paragraph{Historical Bias ($\Delta_{S}$):\footnote{\citet{mehrabi2021} defines \emph{Historical bias} as ``the already existing bias and socio-technical issues in the world'' that ``can seep ... from the data generation process even given a perfect sampling and feature selection.''~\citep{suresh2021framework}.}} denotes the difference in accuracy (F1 score) between stereotypical sentences and anti-stereotypical sentences. Similarly to population bias, $\Delta_S \in [-100, 100]$, and positive values indicate that the model performs better on stereotypical gender role assignments.

\subsection{Results}

\begin{figure}[tb!]
    \centering
    \includegraphics[trim=15 0 0 0,clip,width=\linewidth]{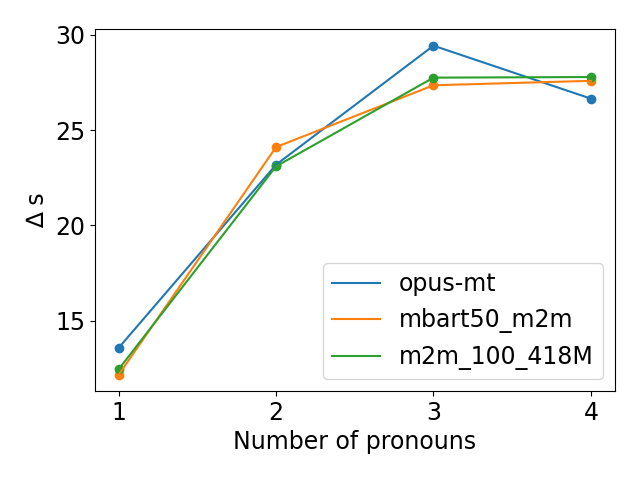}
    
    \caption{Historical gender bias ($\Delta_S$) in machine translation models
    by the number of pronouns in the sentence. Indicating that while the bias is witnessed with a single pronoun, it is exacerbated in sentences with more pronouns.}
    \label{fig:pronouns}
\end{figure}

\begin{figure}[tb!]
    \centering
    \includegraphics[trim=15 0 0 0,clip,width=\linewidth]{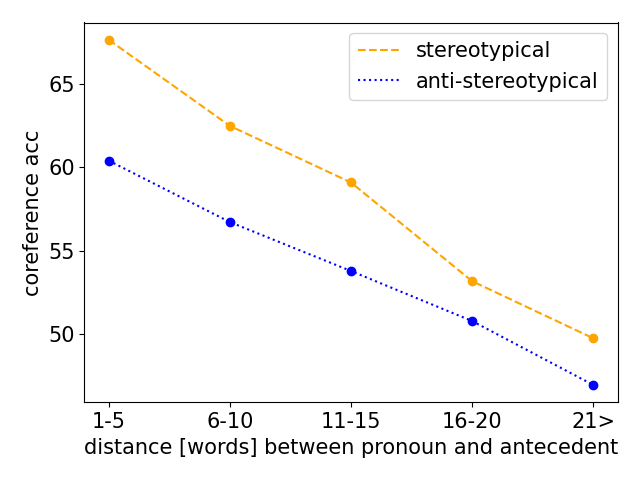}
    
    \caption{Coreference resolution performance as a function of the distance between pronoun and antecedent for stereotypical (orange) and anti-stereotypical (blue). The performance on both partitions deteriorates towards random choice the farther apart the two elements are.}
    \label{fig:dist}
\end{figure}

The results for gender bias in machine translation are presented in Table~\ref{tab:mt-results}, and the results for coreference resolution are presented in the first row in Table~\ref{tab:coref-results}.
We draw various findings and observations based on these results and additional analyses.

\paragraph{All tested models for machine translation and coreference resolution are prone to gender bias on real-world texts.}
Both $\Delta_G$ and $\Delta_S$ are larger than zero across all settings, indicating that all models perform better on entities co-referring with a  masculine pronoun and over-rely on gender stereotypes, even when it is in conflict with the pronouns providing contextual gender indications. To the best of our knowledge, this is the first time this phenomenon was observed and quantified at large scale on real-world instances, especially important for popular NLP services, such as machine translation and coreference resolution, which are in common use in many downstream applications.

\paragraph{Machine translation models do worse on sentences with many pronouns.}
Figure~\ref{fig:pronouns} breaks down $\Delta_S$ for machine translation as function of the number of pronouns in the sentence, showing that machine translation models are prone to fallback to their stereotypes the more pronouns appear in the sentence. This may be due to the increased syntactic complexity presented in such sentences.

\paragraph{Coreference resolution performance deteriorates towards random choice the longer the distance between pronoun and antecedent.}
Figure~\ref{fig:dist} shows that the larger the distance (in words) between entity and coreferring pronoun, SpanBERT's performance deteriorates towards random choice, for both stereotypical and anti-stereotypical partitions, diminishing the difference in performance between them. 

\paragraph{Performance varies across domains.}
We compare gender bias across each of BUG's three domains in Figure~\ref{fig:main}.
It seems that m2m\_100\_418M is the noisiest model in terms of gender bias, its accuracy is the lowest among all languages except Hebrew, and its $\Delta_G$ is the highest. In contrast, mBART50\_m2m is the most accurate model among the three on all languages except Spanish and Hebrew, and its $\Delta_G$ is the lowest on all languages except Arabic and German. A possible explanation may be the vast difference in number of training parameters (15B for mBART50\_m2m versus 418M in m2m\_100\_418M). 
Notably, m2m\_100\_418M achieves a negative $\Delta_S$ score on PubMed~(Figure~\ref{fig:main}), indicating that it over translates entities using \emph{anti-stereotypical} inflections (e.g., preferring to translate engineers as female).
However, the model's low accuracy and high $\Delta_G$ score on the same corpus may indicate that this is mostly due to a noisy translation output, perhaps due to the scientific domain of the input texts in PubMed.

\paragraph{Our findings support previous work.}
The accuracy of the translations in this evaluation are much higher than that found by \citet{stanovsky-etal-2019-evaluating} and \citet{zhao-etal-2018-gender} work~(69.9\% in average vs. 47.6\%), because of BUG's 3:1 ratio in favor of masculine entities versus feminine entities and 2:1 ratio in favor of stereotypical sentences versus anti-stereotypical sentences, representing a distribution which is closer to real-world use-cases. However, $\Delta_G$ and $\Delta_S$ are relative and their values are similar to those found in previous work, indicating that in fact all tested models were prone to gender bias.
In addition, we find that all machine translation models achieve best performance on Czech as a target language, corroborating the findings of \citet{kocmi-etal-2020-gender}, and that Russian and Hebrew have the highest $\Delta_G$ and $\Delta_S$ respectively, again confirming previous findings~\citep{stanovsky-etal-2019-evaluating}. 
For coreference resolution, SpanBERT's gender bias $\Delta_S$ metric in Table~\ref{tab:coref-results} is better (i.e., smaller) than the models reported by~\citep{zhao-etal-2018-gender} (6.0 versus 13.5), which again may be due to the increase in number of parameters.

\begin{table}[]
\centering
\resizebox{\columnwidth}{!}{%
\begin{tabular}{@{}lccc@{}}
\toprule
Coreference Model     & Acc           & $\Delta_G$   & $\Delta_S$   \\ \midrule
SpanBERT        & \textbf{65.1} & 10.2         & 6.0          \\
SpanBERT + {\small anti-stereotypical BUG} & 64.1          & \textbf{5.8} & \textbf{2.9} \\ \bottomrule
\end{tabular}}
\caption{Results for gender bias in coreference resolution. The first row indicates the performance of off-the-shelf SpanBERT on our human validated annotations (Gold BUG), showing that it tends to overperform when clustering masculine and stereotypical gender role assignments. The second row depicts results after finetuning on the anti-stereotypical portion of BUG, showing a 50\% error reduction at the cost of a 1\% absolute reduction in accuracy.}
\label{tab:coref-results}
\end{table}

\section{Debiasing with BUG}
\label{sec:debiasing}
Finally, we show that BUG's size and diverse instances make it amenable for finetuning, which results in more robust models, less prone to rely on gender stereotypes.

In the second row in Table~\ref{tab:coref-results} we report results of finetuning
SpanBERT on the anti-stereotypical portion of BUG (consisting of 29.9K instances), and reevaluate its gender bias metrics on the held out human validated instances (Gold BUG, 1,720 instances).
The motivation is to overexpose the coreference model to anti-stereotypical gender role assignment, where relying on stereotypes would directly hurt performance.
Indeed, this yields a relative error reduction of more than 50\% (3\% absolute improvement).

We note however, that this comes at the cost of an absolute 1\% drop in overall performance accuracy, which may be an expected side-effect due to the shift in training set distribution.
Future work can explore ways to find better trade-offs between accuracy and reliance of gender bias with the help of BUG.

\section{Related work}
\label{sec:related_work}
Several works created synthetic datasets to evaluate gender bias 
 \citep{kiritchenko-mohammad-2018-examining, gonzalez-etal-2020-type, Renduchintala2021InvestigatingFO}, e.g.,  in the context of coreference \citep{rudinger-etal-2017-social, zhao-etal-2018-gender} and machine translation \citep{stanovsky-etal-2019-evaluating, Prates2019AssessingGB, kocmi-etal-2020-gender}, and some works used synthetic datasets to debias models \citep{saunders-etal-2020-neural, zhao-etal-2018-gender}.
 
\citet{webster-etal-2018-mind} and \citet{gonen-webster-2020-automatically}, collected natural  medium-scale (4.4K sentences) datasets from Wikipedia and reddit, respectively, and use them to evaluate gender bias in models of coreference resolution and machine translation. However, their datasets focused on the difference in performance between masculine and feminine entities (population bias), while in this work we also measure historical bias as the difference in performance between stereotypical and anti-stereotypical gender role assignment. In Section~\ref{sec:data}, we compare BUG to these datasets, finding it is more diverse and challenging in various respects.

\section{Conclusion and Future Work}
We presented  BUG, a large-scale corpus of 108K diverse real-world English sentences, collected via semi-automatic grammatical pattern matching.
We use BUG to evaluate gender bias in various coreference resolution and machine translation models, finding that models tend to make predictions in accordance with gender stereotypes, even when in conflict with opposite gendered pronouns in the sentence.
Finally, we finetuned a coreference resolution model on BUG, finding it reduces its gender bias on a held out set.
Our data and code are publicly available at \url{\repolink}.



Future work can extend BUG by including more patterns and by extracting sentences from  corpora with gold annotations for machine translation and coreference resolution. This will allow exploration of the effect that exposure to anti-stereotypical examples during finetuning has on gender bias reduction.

\section*{Acknowledgements}
We thank \spiketeam{} for their help with 
SPIKE during our experiments, for fruitful discussions and their comments on earlier drafts of the paper, and 
the anonymous reviewers for their helpful comments
and feedback.  This work was supported in part by  a research gift from the Allen Institute for AI.


\bibliography{custom}
\bibliographystyle{acl_natbib}

\end{document}